\documentclass[11pt,a4paper,table,dvipsnames]{article}
\usepackage[hyperref]{tlt16}

\usepackage[english]{babel}
\usepackage[utf8]{inputenc}
\usepackage[T1]{fontenc}
\usepackage{times}

\usepackage{amsmath,amssymb,amsthm,amsfonts}

\usepackage{tabularx}
\usepackage{booktabs}
\usepackage{graphicx}
\usepackage{subfigure}

\usepackage{relsize}
\usepackage[smaller,nolist]{acronym}
\usepackage{url}

\usepackage{soul}
\usepackage[subtle]{savetrees}

\aclfinalcopy 

\graphicspath{{images/}}
\DeclareGraphicsExtensions{.eps,.pdf,.png,.jpg}
\newcommand{\ontonotes}[0]{\emph{OntoNotes 5.0}}
\newcommand{\glove}[0]{\emph{Glove}}
\newcommand{\tensorflow}[0]{\emph{TensorFlow}}
\newcommand{\spacy}[0]{\emph{Spacy v1.8.2}}

\newcolumntype{b}{>{\arraybackslash}X}
\newcolumntype{m}{>{\centering\arraybackslash\hsize=.25\hsize}X}
\newcolumntype{s}{>{\centering\arraybackslash\hsize=.125\hsize}X}

\title{Graph Convolutional Networks for Named Entity Recognition}

\author{
    Cetoli,~A. \quad Bragaglia,~S. \quad O'Harney,~A.\,D. \quad Sloan,~M.\\
    Context Scout\\
    \texttt{\{alberto, stefano, andy, marc\}@contextscout.com}
}

\begin{document}

\begin{abstract}
In this paper we investigate the role of the dependency tree in a named entity recognizer upon using a set of \acp{GCN}. 
We perform a comparison among different \ac{NER} architectures and show that the grammar of a sentence positively influences the results. 
Experiments on the \ontonotes{} dataset demonstrate consistent performance improvements, without requiring heavy feature engineering nor additional language\--specific knowledge~\footnote{A version of this system can be found at \url{https://github.com/contextscout/gcn_ner}.}.
\end{abstract}

\maketitle

\section{Introduction and Motivations} 
\label{sec:introduction_and_motivations}

The recent article by \citeauthor{MarcheggianiT17}~\cite{MarcheggianiT17} opened the way for a novel method in \ac{NLP}. In their work, they adopt a \ac{GCN}~\cite{KipfW16} approach to perform semantic role labeling, improving upon previous architectures.
While their article is specific to recognizing the predicate-argument structure of a sentence, their method can be applied in other areas of \ac{NLP}. 
One example is \ac{NER}.

High performing statistical approaches have been used in the past for entity recognition, notably \textit{Markov} models~\cite{McCallum2000}, \acp{CRF}~\cite{Lafferty2001}, and \acp{SVM}~\cite{Takeuchi2002}. 
More recently, the use of neural networks has become common in \ac{NER}.

The method proposed by \citeauthor{CollobertAA11}~\cite{CollobertAA11} suggests that a simple feed\--forward network can produce competitive results with respect to other approaches. 
Shortly thereafter, \citeauthor{ChiuN15}~\cite{ChiuN15} employed \acp{RNN} to address the problem of entity recognition, thus achieving state of the art results. 
Their key improvements were twofold: using a bi\--directional \ac{LSTM} in place of a feed\--forward network and concatenating morphological information to the input vectors.

Subsequently, various improvements appeared: using a \ac{CRF} as a last layer~\cite{HuangXY15} in place of a \texttt{softmax} function, a gated approach to concatenating morphology~\cite{CaoR16} and predicting nearby words~\cite{Rei17}. 
All such methods, however, understand text as a one dimensional collection of input vectors; any syntactic information -- namely the parse tree of the sentence -- is ignored.

We believe that dependency trees and other linguistic features play a key role on the accuracy of \ac{NER} and that \acp{GCN} can grant the flexibility and convenience of use that we desire.
In this paper our contribution is twofold: on one hand, we introduce a methodology for tackling entity recognition with \acp{GCN}; 
on the other hand we measure the impact of using dependency trees for entity classification upon comparing the results with prior solutions. At this stage our goal is not to beat the state-of-the-art but rather to quantify the effect of our novel architecture.

The remainder of the article is organized as follows: in Section \ref{sec:methods_and_materials} we introduce the theoretical framework for our methodology, then the features considered in our model and eventually the training details. 
Section \ref{sec:experimental_results} describes the experiments and presents the results. We discuss relevant works in Section \ref{sec:related_works} and draw the conclusions in Section \ref{sec:concluding_remarks}.

\section{Methods and Materials} 
\label{sec:methods_and_materials}
\subsection{Theoretical Aspects} 
\label{sub:theoretical_aspects}
\begin{figure}[t!]
    \centering
        \includegraphics[width=0.9\textwidth]{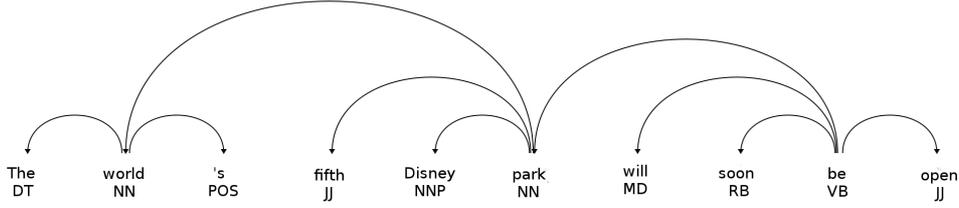}   
    \caption{An example sentence along with its dependency graph. \acp{GCN} propagate the information of a node to its nearest neighbours.}
    \label{fig:sentence}
\end{figure}

\aclp{GCN}~\cite{KipfW16} operate on graphs by convolving the features of neighbouring nodes. 
A \ac{GCN} layer propagates the information of a node onto its nearest neighbours. 
By stacking together $N$ layers, the network can propagate the features of nodes that are at most $N$ hops away.

While the original formulation did not include directed graphs, they were further extended in \cite{MarcheggianiT17} to be used on directed syntactic/dependency trees. 
In the following we rely on \citeauthor{MarcheggianiT17}'s work to assemble our network.

Each \ac{GCN} layer creates new node embeddings by using neighbouring nodes and these layers can be stacked upon each other. 
In the undirected graph case, the information at the $k^{st}$ layer is propagated to the next one according to the equation

\begin{equation}
    h_v^{k+1} = \mathrm{ReLU} \left(
        \sum_{u \in \mathcal{N}(v)} \left( 
            W^{k}h_u^{k} + b^{k} 
        \right) 
    \right) \,,
    \label{eq:undirected_gcn}
\end{equation}

\noindent where $u$ and $v$ are nodes in the graph. 
$\mathcal{N}$ is the set of nearest neighbours of node $v$, plus the node $v$ itself.
The vector $h_u^{k}$ represents node $u$'s embeddings at the $k^\mathrm{st}$ layer, while $W$ and $b$ are are a weight matrix and a bias -- learned during training -- that map the embeddings of node $u$ onto the adjacent nodes in the graph; $h_u$ belongs to $\mathbb{R}^{m}$, $W \in \mathbb{R}^{m \times m}$ and $b \in \mathbb{R}^{m}$.

Following the example in \cite{MarcheggianiT17}, we prefer to exploit the directness of the graph in our system. 
Our inspiration comes from the bi\--directional architecture of stacked \acp{RNN}, where two different neural networks operate forward and backward
respectively. 
Eventually the output of the \acp{RNN} is concatenated and passed to further layers.

In our architecture we employ two stacked \acp{GCN}: One that only considers the incoming edges for each node

\begin{equation}
    \overleftarrow{h}_v^{k+1} = \mathrm{ReLU} \left(
        \sum_{u \in \mathcal{\overleftarrow{N}}(v)} \left( 
            \overleftarrow{W}^{k} h_u^{k} + \overleftarrow{b}^{k} 
        \right)
    \right) \,,
\end{equation}

\noindent and one that considers only the outgoing edges from each node

\begin{equation}
    \overrightarrow{h}_v^{k+1} = \mathrm{ReLU} \left(
        \sum_{u \in \mathcal{\overrightarrow{N}}(v)} \left( 
            \overrightarrow{W}^{k}h_u^{k} + \overrightarrow{b}^{k} 
        \right)
    \right) \,.
\end{equation}

After $N$ layers the final output of the two \acp{GCN} is the concatenation of the two separated layers

\begin{equation}
    h_v^N = \overrightarrow{h}_v^{N} \oplus \overleftarrow{h}_v^{N} \,.
    \label{eq:bidir_gcn}
\end{equation}

In the following, we refer to the architecture expressed by \equationautorefname~\ref{eq:bidir_gcn} as a \emph{\textbf{bi\--directional \ac{GCN}}}.

\subsection{Implementations Details} 
\label{sub:implementation_details}
\subsubsection{Models} 
\label{subs:models}
\begin{figure}[t!]
    \centering
        \subfigure[][]{%
            \label{fig:lstm}%
            \includegraphics[width=0.4\textwidth]{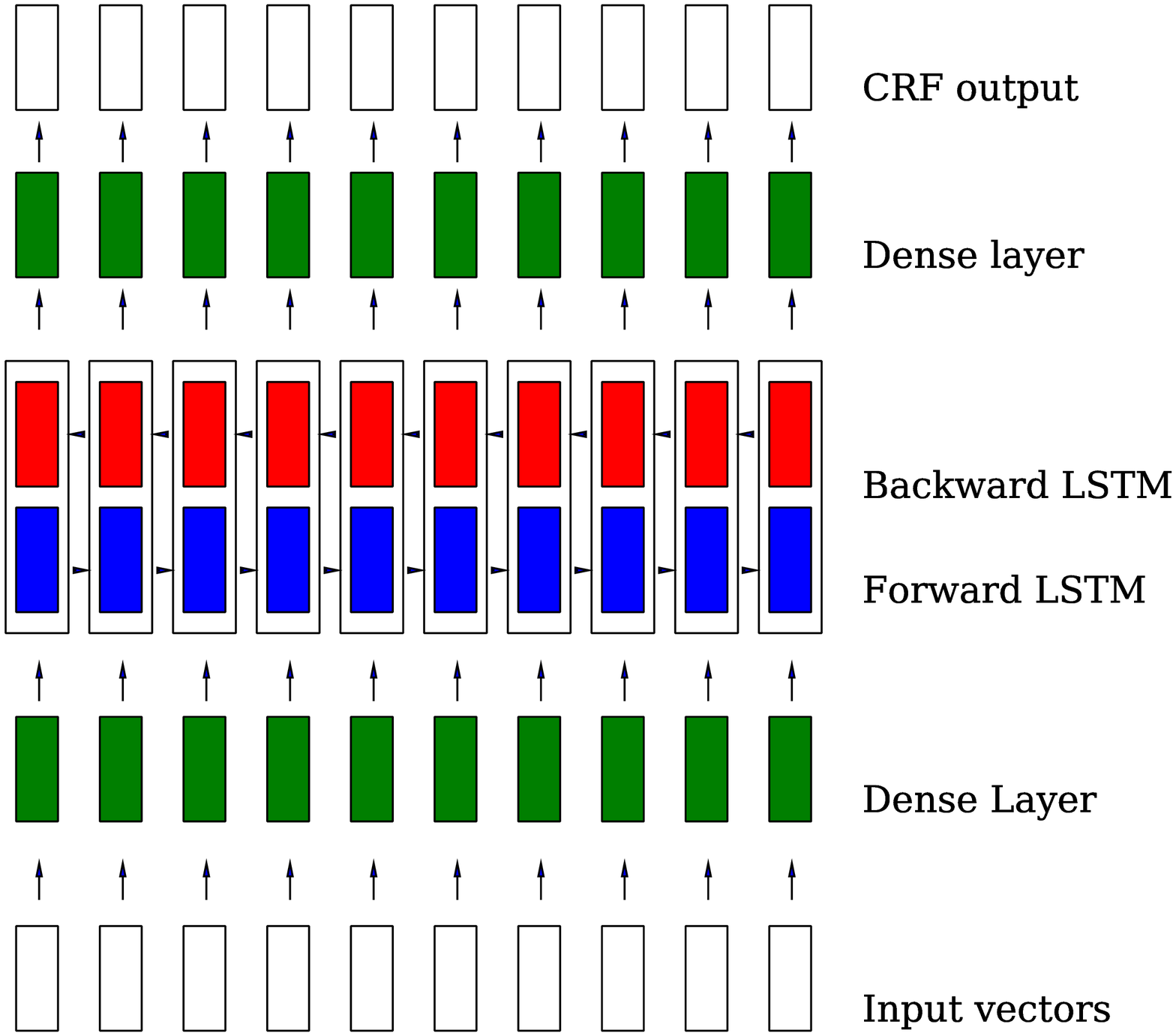}}%
        \qquad
        \subfigure[][]{%
            \label{fig:gcn}%
            \includegraphics[width=0.4\textwidth]{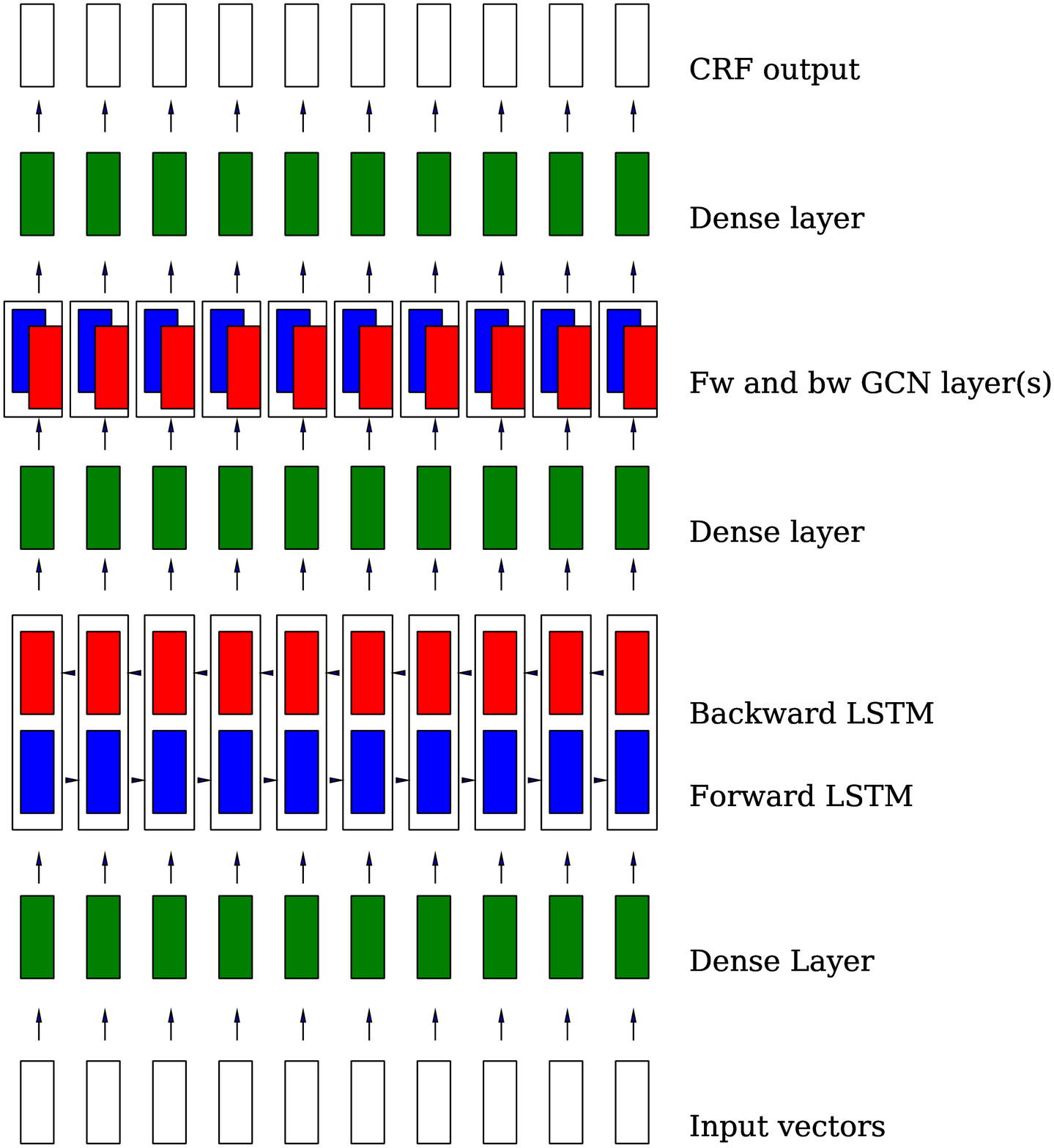}}%
    \caption{bi\--directional architectures:
        \subref{fig:lstm} \ac{LSTM}; and, \subref{fig:gcn} \ac{GCN} layers.}
    \label{fig:lstm_gcn} 
\end{figure}

Our architecture is inspired by the work of \citeauthor{ChiuN15}~\cite{ChiuN15}, \citeauthor{HuangXY15}~\cite{HuangXY15}, and \citeauthor{MarcheggianiT17}~\cite{MarcheggianiT17}. 
We aim to combine a \ac{Bi-LSTM} model with \acp{GCN}, using \ac{CRF} as the last layer in place of a \texttt{softmax} function.

We employ seven different configurations by selecting from two sets of \ac{PoS} tags and two sets of word embedding vectors. 
All the models share a bi\--directional \ac{LSTM} which acts as the foundation upon which we apply our \ac{GCN}.
The different combinations are built using the following elements:

\paragraph{Bi-LSTM} 
\label{par:bi_lstm}
We use a bi\--directional \ac{LSTM} structured as in \figurename~\ref{fig:lstm}. 
The output is mediated by two fully connected layers ending in a \ac{CRF}~\cite{HuangXY15}, modelled as a Viterbi sequence.
The best results in the \emph{dev set} of \ontonotes{} were obtained upon staking two \ac{LSTM} layers, both for the forward and backward configuration. 
This is the number of layers we keep in the rest of our work.
This configuration -- when used alone -- is a consistency test with respect to the previous works. 
As seen in \tablename~\ref{tab:results}, our findings are compatible with the results in \cite{ChiuN15}.

\paragraph{Bi-GCN} 
\label{par:bi_gcn}
In this model, we use the architecture created in \cite{MarcheggianiT17} where a \ac{GCN} is applied on top of a \ac{Bi-LSTM}. 
This system is shown in \figurename~\ref{fig:gcn} (right side). 
The parser used to create the dependency tree is \spacy{}~\cite{HonnibalJ2015}.
The best results in the \emph{dev set} were obtained upon using only one \ac{GCN} layer, and we use this configuration through our models.
We employ two different embedding vectors for this configuration: one in which only word embeddings are fed as an input, the other one where
\ac{PoS} tag embeddings are concatenated to the word vectors.

\paragraph{Input vectors} 
\label{par:input_vectors}
We use three sets of input vectors. 
First, we simply employ the word embeddings found in the \glove{} vectors~\cite{pennington2014glove}:

\begin{equation}
    x_{\mathrm{input}} = x_{\mathrm{glove}}.
\end{equation}

\noindent In the following, we employ the $300$ dimensional vector from two different distributions: one with $1$M words and another one with $2.2$M words. 
Whenever a word is not present in the \glove{} vocabulary we use the vector corresponding to the word ``entity'' instead.

The second type of vector embeddings concatenates the \glove{} word vectors with \ac{PoS} tags embeddings. 
We use randomly initialized \acl{PoS} embeddings that are allowed to fine-tune during training:

\begin{equation}
    x_{\mathrm{input}} = x_{\mathrm{glove}} \oplus x_{\mathrm{PoS}}.
    \label{eq:word_pos_input}
\end{equation}

The final quality of our results correlates to the quality of our \acl{PoS} tagging. 
In one batch we use the manually curated \ac{PoS} tags included in the \ontonotes{} dataset~\cite{Ontonotes5} (\emph{\ac{PoS} (gold)}). 
These tags have the highest quality.

In another batch, we use the \ac{PoS} tagging inferred from the parser (\emph{\ac{PoS} (inferred)}) instead of using the manually tagged ones. 
These \ac{PoS} tags are of lower quality.
An external tagger might provide a different number of tokens compared to the ones present in the training and evaluation datasets. 
This presents a challenge.
We skip these sentences during training, while considering the entities in such sentences as incorrectly tagged during evaluation.

Finally, we add the morphological information to the feature vector for the third type of word embeddings. 
The reason -- explained in \cite{CaoR16} -- is that out\--of\--vocabulary words are handled badly whilst using only word embeddings:

\begin{equation}
    x_{\mathrm{input}} = x_{\mathrm{glove}} \oplus x_{\mathrm{PoS}} \oplus x_{\mathrm{morphology}}.
\end{equation}

\noindent We employ a bi\--directional \ac{RNN} to encode character information. 
The end nodes of the \ac{RNN} are concatenated and passed to a dense layer, which is integrated to the feature vector along with the embeddings
and \ac{PoS} information.
In order to speed up the computation, we truncate the words by keeping only the first 12 characters. 
Truncation is not commonly done, as it hinders the network's performance; we leave further analysis to following works.

\begin{figure}[t!]
    \centering
        \includegraphics[width=0.7\textwidth]{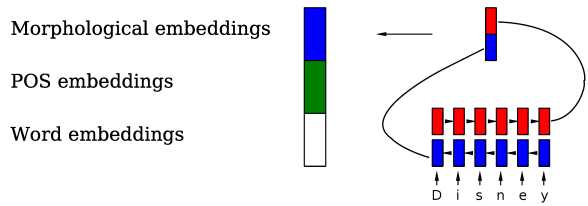} 
    \caption{Feature vector components. Our input vectors have up to three components: the word embeddings, the \ac{PoS} embeddings, and a morphological embedding obtained through feeding each word to a \ac{Bi-LSTM} and then concatenating the first and last hidden state.}
    \label{fig:vectors}
\end{figure}

\paragraph{Dropout} 
\label{par:dropout}
In order to tackle over\--fitting, we apply dropout to all the layers on top of the \ac{LSTM}.  
The probability to drop a node is set at $20\%$ for all the configurations. 
The layers that are used as input to the \ac{LSTM} do not use dropout.

\paragraph{Network output} 
\label{par:network_output}
At inference time, the output of the network is a $19$-dimensional vector for each input word. 
This dimensionality comes from the $18$ tags used in \ontonotes{}, with an additional dimension which expresses the absence of a named entity.
No \ac{BIOES} markings are applied; at evaluation time we simply consider a \emph{name chunk} as a contiguous sequence of words belonging to the same category.

\subsubsection{Training} 
\label{subs:training}
We use \tensorflow{}~\cite{tensorflow2015-whitepaper} to implement our neural network.  
Training and inference is done at the sentence level.
The weights are initialized randomly from the uniform distribution and the initial state of the \acp{LSTM} are set to zero.  
The system uses the configuration in \appendixname~\ref{app:configuration}.

The training function is the \ac{CRF} loss function as explained in \cite{HuangXY15}. 
Following their notation, we define $[f]_{i,t}$ as the matrix that represents the score of the network for the $t^{th}$ word to have the $i^{th}$ tag.
We also introduce $A_{ij}$ as the transition matrix which stores the probability of going from tag $i$ to tag $j$. 
The transition matrix is usually trained along with the other network weights. 
In our work we preferred instead to set it as constant and equal to the transition frequencies as found in the training dataset.

The function $f$ is an argument of the network's parameters $\theta$ and the input sentence $[x]_1^T$ (the list of embeddings with length $T$).
Let the list of $T$ training labels be written as $[i]_1^T$, then our loss function is written as

\begin{equation}
    \mathcal{S}\left(
        [x]_1^T,[i]_1^T,\theta,A_{ij}
    \right) - \sum_{[j]_1^T} \mathrm{exp}\left(
        [x]_1^T,[j]_1^T,\theta,A_{ij} + [f]_{[i]_t,t}
    \right) \,,
\end{equation}

\noindent where

\begin{equation}
    \mathcal{S}\left(
        [x]_1^T,[i]_1^T,\theta,A_{ij}
    \right) = \sum_1^T\left( 
        A_{[i]_{t-1},[i]_t} + f(\theta,A_{ij})
    \right) \,.
\end{equation}

At inference time, we rely on the Viterbi algorithm to find the sequence of tokens that maximizes $\mathcal{S}\left([x]_1^T,[i]_1^T,\theta,A_{ij}\right)$.
We apply mini\--batch stochastic gradient descent with the \emph{Adam} optimiser~\cite{KingmaB14}, using a learning rate fixed to $10^{-4}$.

\section{Experimental Results} 
\label{sec:experimental_results}
In this section, we compare the different methods applied and discuss the results.
The scores in \tablename~\ref{tab:results}~\footnote{The results from \citeauthor{RatinovRo09} and \citeauthor{Finkel09} are taken from \cite{ChiuN15}.} are presented as an average of $6$ runs with the error being the standard deviation; we keep only the first significant digit of the errors, approximating to the nearest number.

\begin{table}[ht!]
    \centering
    \begin{tabularx}{\textwidth}{|b||s|s|m||s|s|m|}
        \toprule
              & \multicolumn{3}{c||}{\textsc{\textbf{Dev}}} & \multicolumn{3}{c|}{\textsc{\textbf{Test}}} \\
            \multicolumn{1}{|c||}{\textbf{Description}} & \textbf{prec} & \textbf{rec} & \textbf{$F_1$} & \textbf{prec} & \textbf{rec} & \textbf{$F_1$} \\
        \midrule
            {\tiny \ac{Bi-LSTM} + $1$M \glove{} + \ac{CRF}}                                                 & $80.9$ & $78.2$ & $79.5 \pm 0.3$ & $79.1$ & $75.9$ & $77.5 \pm 0.4$ \\
        \hline
            {\tiny \ac{Bi-LSTM} + $1$M \glove{} + \ac{CRF} + \ac{GCN}}                                      & $82.2$ & $79.5$ & $80.8 \pm 0.3$ & $82.0$ & $77.5$ & $79.7 \pm 0.3$ \\
            {\tiny \ac{Bi-LSTM} + $1$M \glove{} + \ac{CRF} + \ac{GCN} + \ac{PoS} (gold)}                    & $82.1$ & $83.7$ & $82.9 \pm 0.3$ & $82.4$ & $81.8$ & $82.1 \pm 0.4$ \\
            {\tiny \ac{Bi-LSTM} + $2.2$M \glove{} + \ac{CRF} + \ac{GCN} + \ac{PoS} (gold)}                  & $83.3$ & $84.1$ & $83.7 \pm 0.4$ & $83.6$ & $82.1$ & $82.8 \pm 0.3$\\
            {\tiny \ac{Bi-LSTM} + $2.2$M \glove{} + \ac{CRF} + \ac{GCN} + \ac{PoS} (inferred)}              & $83.8$ & $82.9$ & $83.4 \pm 0.4$ & $82.2$ & $80.5$ & $81.4 \pm 0.3$\\
            {\tiny \ac{Bi-LSTM} + $2.2$M \glove{} + \ac{CRF} + \ac{GCN} + \ac{PoS} (gold) + Morphology}     & $86.6$ & $82.7$ & $84.6 \pm 0.4$ & $86.7$ & $80.7$ & $83.6 \pm 0.4$\\
            {\tiny \ac{Bi-LSTM} + $2.2$M \glove{} + \ac{CRF} + \ac{GCN} + \ac{PoS} (inferred) + Morphology} & $85.3$ & $82.3$ & $83.8 \pm 0.4$ & $84.3$ & $80.1$ & $82.0 \pm 0.4$ \\
        \midrule
            {\tiny \citeauthor{ChiuN15}~\cite{ChiuN15}}                                                & & & $\textbf{84.6} \pm 0.3$ & $86.0$ & $86.5$ & $\textbf{86.3} \pm 0.3$ \\
            {\tiny \citeauthor{RatinovRo09}~\cite{RatinovRo09}}                                        & & & & $82.0$ & $84.9$ & $83.4 \textcolor{white}{ \pm 0.0}$ \\
            {\tiny \citeauthor{Finkel09}~\cite{Finkel09}}                                              & & & & $84.0$ & $80.9$ & $82.4 \textcolor{white}{ \pm 0.0}$ \\
            {\tiny \citeauthor{DurrettK15}~\cite{DurrettK15}}                                          & & & & $85.2$ & $82.9$ & $84.0 \textcolor{white}{ \pm 0.0}$ \\
        \bottomrule
    \end{tabularx}
    \caption{Results of our architecture compared to previous findings.}
    \label{tab:results} 
\end{table}

The results show an improvement of $2.2 \pm 0.5\%$ upon using a \ac{GCN}, compared to the baseline result of a bi\--directional \ac{LSTM} alone ($1^{st}$ row). 
When concatenating the gold \ac{PoS} tag embedding in the input vectors, this improvement raises to $4.6 \pm 0.6\%$.
However, gold \ac{PoS} tags cannot be used outside the \ontonotes{} dataset. 
The $F_1$ score improvement for the system while using inferred tags (from the parser) is lower: $3.2 \pm 0.6\%$.

For comparison, increasing the size of the \glove{} vector from $1$M to $2.2$M gave an improvement of $0.7 \pm 0.5\%$. 
Adding the morphological information of the words, albeit truncated at 12 characters, improves the $F_1$ score by $2.2 \pm 0.5\%$.

Our results strongly suggest that syntactic information is relevant in capturing the role of a word in a sentence, and understanding sentences as one\--dimensional lists of words appears as a partial approach. 
Sentences embed meaning through internal graph structures: the graph convolutional method approach -- used in conjunction with a parser (or a \emph{treebank}) -- seems to provide a lightweight architecture that incorporates grammar while extracting named entities.

Our results -- while competitive -- fall short of achieving the state of the art. 
We believe this to be the result of a few factors: we do not employ \ac{BIOES} annotations for our tags, lexicon and capitalisation features are ignored, and we truncate words when encoding the morphological vectors.
Our main claim is nonetheless clear: grammatical information positively boosts the performance of recognizing entities, leaving further improvements to be explored.

\section{Related Works} 
\label{sec:related_works}
There is a large corpus of work on named entity recognition, with few studies using explicitly non\--local information for the task. 
One early work by \citeauthor{Finkel05}~\cite{Finkel05} uses Gibbs sampling to capture long distance structures that are common in language use. 
Another article by the same authors uses a joint representation for constituency parsing and \ac{NER}, improving both techniques. 
In addition, dependency structures have also been used to boost the recognition of bio\--medical events \cite{McClosky11} and for automatic content extraction \cite{Li13}.

Recently, there has been a significant effort to improve the accuracy of classifiers by going beyond vector representation for sentences. 
Notably the work of \citeauthor{Peng17}~\cite{Peng17} introduces \emph{graph LSTMs} to encode the meaning of a sentence by using dependency graphs. 
Similarly \citeauthor{DhingraYCS17}~\cite{DhingraYCS17} employ \emph{\acp{GRU}} that encode the information of acyclic graphs to achieve state\--of\--the\--art results in co-reference resolution.

\section{Concluding Remarks} 
\label{sec:concluding_remarks}
We showed that dependency trees play a positive role for entity recognition by using a \ac{GCN} to boost the results of a bidirectional \ac{LSTM}. In addition, we modified the standard convolutional network architecture and introduced a bidirectional mechanism for convolving directed graphs. This model is able to improve upon the \ac{LSTM} baseline:
Our best result yielded an improvement of $4.6 \pm 0.6\%$ in the $F_1$ score, using a combination of both \ac{GCN} and \ac{PoS} tag embeddings.

Finally, we prove that \acp{GCN} can be used in conjunction with different techniques. 
We have shown that morphological information in the input vectors does not conflict with graph convolutions. 
Additional techniques, such as the gating of the components of input vectors~\cite{ReiCP16} or neighbouring word prediction~\cite{Rei17} should be
tested together with \acp{GCN}. 
We will investigate those results in future works.

\bibliographystyle{acl_natbib}
\bibliography{articles}

\clearpage

\appendix

\section{Configuration} 
\label{app:configuration}
\begin{table}[h!]
    \centering
    \begin{tabularx}{0.7\textwidth}{|X|r|}
        \toprule
            \multicolumn{1}{|c}{\textbf{Parameter}} & \multicolumn{1}{|c|}{\textbf{Value}} \\
        \midrule
            \glove{} word embeddings & $300$ dim \\
            \ac{PoS} embedding       & $15$ dim \\
            Morphological embedding  & $20$ dim \\ 
        \midrule
            First dense layer        & $40$ dim \\
            \ac{LSTM} memory         & ($2\times$) $160$ dim \\
            Second dense layer       & $160$ dim \\
        \midrule
            \ac{GCN} layer           & ($2\times$) $160$ dim \\
            Final dense layer        & $160$ dim \\
        \midrule
            Output layer             & $16$ dim \\
        \midrule
            Dropout                   & $0.8$ (\emph{keep probability}) \\
        \bottomrule
    \end{tabularx}
    \caption{Summary of the configuration used for training the network.}
    \label{tab:configuration}
\end{table}
    
\begin{acronym}[LSTM]
    \acro{BIOES}{Begin, Inside, Outside, End, Single}
    \acro{Bi-GCN}{Bi\--directional Graph Convolutional Network}
    \acro{Bi-LSTM}{Bi\--directional Long Short-Term Memory}
    \acro{CRF}{Conditional Random Field}
    \acro{GCN}{Graph Convolutional Network}
    \acro{GRU}{Gated Recurrent Unit}
    \acro{LSTM}{Long Short-Term Memory}
    \acro{NER}{Named Entity Recognition}
    \acro{NLP}{Natural Language Processing}
    \acro{PoS}{Part\--of\--Speech}
    \acro{RNN}{Recurrent Neural Network}
    \acro{SVM}{Support Vector Machine}
\end{acronym}

\end{document}